\documentclass[sigconf,nonacm]{acmart}
\AtBeginDocument{%
  }

\usepackage{multicol}
\usepackage{multirow}

\usepackage{algorithmicx}
\usepackage{algorithm}
\usepackage{algpseudocode}

\usepackage{subcaption}

\begin{document}

\title{NEXICA: Discovering Road Traffic Causality\\(Extended arXiv Version)}

\author{Siddharth Srikanth}
\email{ssrikant@usc.edu}
\affiliation{
    \institution{University of Southern California}
    \city{Los Angeles}
    \state{California}
    \country{USA}
}

\author{John Krumm}
\email{jkrumm@usc.edu}
\affiliation{
    \institution{University of Southern California}
    \city{Los Angeles}
    \state{California}
    \country{USA}
}

\author{Jonathan Qin}
\email{qinjonat@usc.edu}
\affiliation{
    \institution{University of Southern California}
    \city{Los Angeles}
    \state{California}
    \country{USA}
}

\renewcommand{\shortauthors}{Srikanth, Krumm, Qin}

\begin{abstract}
Road traffic congestion is a persistent problem. Focusing resources on the causes of congestion is a potentially efficient strategy for reducing slowdowns. We present \textbf{NEXICA}, an algorithm to discover which parts of the highway system tend to cause slowdowns on other parts of the highway. We use time series of road speeds as inputs to our causal discovery algorithm. Finding other algorithms inadequate, we develop a new approach that is novel in three ways. First, it concentrates on just the presence or absence of events in the time series, where an event indicates the temporal beginning of a traffic slowdown. Second, we develop a probabilistic model using maximum likelihood estimation to compute the probabilities of spontaneous and caused slowdowns between two locations on the highway. Third, we train a binary classifier to identify pairs of cause/effect locations trained on pairs of road locations where we are reasonably certain \textit{a priori} of their causal connections, both positive and negative. We test our approach on six months of road speed data from 195 different highway speed sensors in the Los Angeles area, showing that our approach is superior to state-of-the-art baselines in both accuracy and computation speed.
\end{abstract}




\maketitle

\section{Introduction} \label{sec:introduction}





Traffic congestion cost U.S. drivers an average of 43 hours and a total of \$74 billion in lost time in 2024~\cite{newsweek_traffic_2024}. In 2019, the U.S. freight sector lost \$74.1 billion due to slow traffic~\cite{statista_traffic_2020}. Idling engines also cause more pollution. To reduce these costs, it is important to direct resources efficiently for reducing traffic congestion.

\begin{figure}[t]
    \centering
    \includegraphics[width=1\linewidth]{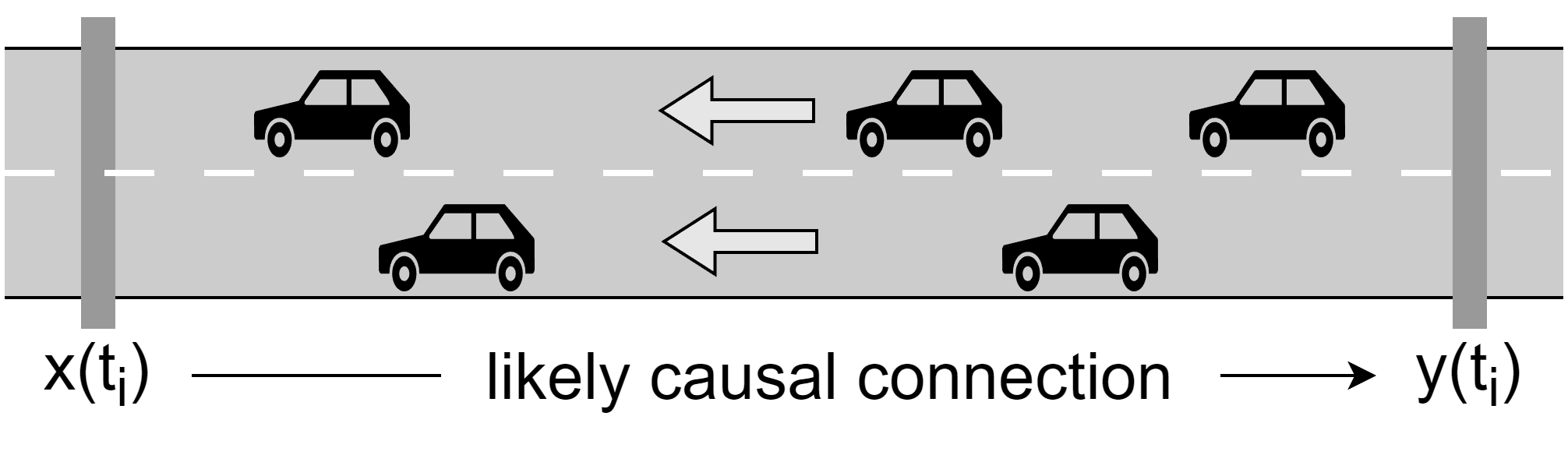}
    \caption{Road speeds at this pair of locations are likely causally connected as a slowdown at $x$ would cause a slowdown at $y$ after some lag in time.}
    \label{fig:ground_truth_positive_example}
\end{figure}

One way to relieve traffic congestion is to restructure the roads, but this is expensive. We need a framework to identify the most effective places to spend. It is not enough to identify the sections of road that consistently suffer from slowdowns, because the congestion may be propagating from other roads. Instead, it is important to identify the causes of the slowdowns, especially those roads that regularly cause slowdowns on other roads. These especially causal roads are good candidates for improvement. While traffic engineers are skilled at identifying causes, the active science of causal discovery, along with detailed data on traffic speeds, can also help to find which parts of the road tend to cause traffic slowdowns. 

Ideally, we would be able to perform explicit experiments to determine causality. In the case of roads, we could intentionally cause a traffic slowdown and observe how it affects traffic speeds on other roads. Since this is impractical, we depend on measured data from the relevant phenomena, using it as a set of natural experiments. The science of causal discovery uses these measured signals to identify causal relationships. For traffic, we are fortunate to have rich data on traffic from sensors in the roads giving time series of traffic speeds measured at different locations. Causal discovery can show which locations on the roads tend to cause slowdowns at other locations.

The first milestone in the science of causal discovery was Granger causality, which, at its core, examines two time series, $x(t_i)$ and $y(t_i)$~\cite{granger1969investigating}. We can imagine that $y(t_i)$ is the time series of road speeds at one road location, and $x(t_i)$ is the time series of road speeds at another road location, as illustrated in Figure~\ref{fig:ground_truth_positive_example}. The $t_i$ represent discrete, evenly sampled points in time. We are interested in discovering if traffic speeds at the location $x$ ``cause'' traffic speeds at location $y$.

Granger causality models $y(t_i)$ with an autoregressive model that says the value of $y(t_i)$ can be predicted by the $p$ past values of $y(t_i)$, where $p$ is some positive integer:

\begin{equation*}
    y(t_i) = \sum_{j=1}^{p} \alpha_j y(t_{i-j}) + \epsilon_i
\end{equation*}

\noindent The $\alpha_j$ values are real coefficients that can be computed on the time series data with least squares. $\epsilon_i$ represents the error in the estimate of $y(t_i)$. If this equation accurately predicts $y(t_i)$, then the signal is somewhat independent and unlikely dependent on another signal, such as $x(t_i)$.

Causal discovery posits that a signal $x(t_i)$ might cause $y(t_i)$. The test begins with an autoregessive model that says the value of $y(t_i)$ is a linear function of its own past values \emph{and} the past values of $x(t_i)$:

\begin{equation*}
    y(t_i) = \sum_{j=1}^{p} \alpha_j y(t_{i-j}) + \sum_{i=1}^{q} \beta_k x(t_{i-k}) + \epsilon_i
\end{equation*}

\noindent The $\alpha_j$ and $\beta_k$ values are real coefficients that can be computed on the time series data with least squares, and $q$ is a positive integer. A series of statistical tests determine the causality. The intuition is that $x$ causes $y$ if the past values of $x$ significantly help in predicting values of $y$. Our tests showed that Granger causality failed to find obvious causal traffic pairs from recorded speed data. One limitation of Granger causality is that it only tests for a linear relationship between the measured values.

Since Granger's test, causal discovery has matured, but the basic paradigm remains the same: There is likely a causal connection from $x$ to $y$ if $x$ somehow helps explain the evolution of $y$. More modern tests can tolerate nonlinear relationships between the variables, accommodate many more than two variables, and produce a causal graph that represents a cascade of causal relationships propagating through a network of variables.

We are focused on finding causal connections between different pairs of locations on roads using time series of traffic speeds. One rare aspect of our causality problem is that we have pairs of locations where, for some pairs, we are quite certain of the causal connection. Locations on the same road that are near each other, with the same direction of travel, with no feeder nor turnouts between them, are likely causally connected, as illustrated in Figure~\ref{fig:ground_truth_positive_example}. Pairs of points on roads that are far apart are most likely not causally connected. Because we can find many examples of ground truth pairs, both positive and negative, we can apply a machine learned, binary classifier to our problem and precisely measure its performance.

In looking for alternatives to Granger causality, we found that two modern causal discovery techniques, PCMCI~\cite{runge2020discovering} and DYNO-TEARS~\cite{pamfil2020dynotears}, do not perform well for our problem. These failures are likely due to the road speed time series having a complicated, nonlinear connection even between obvious causal pairs. Our new algorithm, \textbf{N}atural \textbf{E}vent e\textbf{X}traction for \textbf{I}nference of \textbf{CA}usality (\textbf{NEXICA}), is novel in three ways. First, we replace the time series of traffic speeds with a derived sequence of binary slowdown events. We introduce a simple method for finding the leading edges of slowdown events, and we use these events to assess causality. This means we do not need to make any assumptions nor create any models of the behavior of the raw time series, e.g. linear or nonlinear. The second innovation is that we derive a maximum likelihood method for explicitly estimating the probability that one sequence of binary events causes another sequence. Third, we apply a machine learning classifier to the problem of causal discovery, because we have ground truth causal and noncausal pairs. In most other causality research, there is a paucity of real-world test data, so it is difficult to anticipate accuracy in a general case.
\section{Related Work} \label{sec:related_work}

In addition to traditional Granger causality, described above, there are more modern approaches to causal discovery on time series, such as  PCMCI~\cite{runge2020discovering} and DYNOTEARS~\cite{pamfil2020dynotears}. PCMCI is based on statistical tests of independence among the time series. Its goal is to build a graph of directed causal connections, where the nodes of the graph are the signals (e.g. traffic measurement stations) and the directed edges indicate a causal relationship from cause to effect. PCMCI is robust to confounding causal factors, such as if two signals have a common cause, then the two signals themselves may appear to have a causal connection. Through statistical tests, PCMCI sorts through dependencies to produce a causal graph. The relationships between the signals may be linear or nonlinear. The authors claim PCMCI is suitable for up to hundreds of time series each comprising hundreds of samples. For our traffic causality problem, we experiment with 195 measurement stations with 5-minute speed samples. With six months of data, each station has over 50,000 samples, which we found exceeds the capacity of PCMCI to handle in a reasonable amount of computation time. One of the advantages of our binary, event-based representation of traffic is efficient processing. PCMCI also lacks the ability to learn from ground truth causal pairs, something that is built in to our approach.

PCMCI represents a statistical approach to causal discovery. Another category of algorithm searches directly for the entire causal graph, so-called structure learning. The directed graph is usually constrained to be acyclical, thus a directed, acyclical graph (DAG). The acyclical property ensures that no signal can be an indirect cause of itself. A significant advance in discovering causal DAGs was the paper from Zheng et al. that introduced a continous optimization criteria to enforce the acyclicity constraint, leading to high quality searches~\cite{zheng2018dags}. This paper spawned a cascade of related work, one of which was the DYNOTEARS algorithm for learning causal DAGs from time series data like ours~\cite{pamfil2020dynotears}.  While this work provides a foundation for us to eliminate certain edges based on prior intuition, a major pitfall of this approach is that it does not support a framework to insert causal edges that we know to exist. The DAG restriction also may not be true for traffic, as it is conceivable that a slowdown may reinforce itself through the road network. DYNOTEARS was also very slow to run on our data.

Previous work has focused specifically on the problem of road traffic causality. Queen and Albers present a dynamic Bayesian network for traffic forecasting that focuses on conditional independencies and causes in the relationships between measured traffic at different locations on the road network~\cite{queen2009intervention}. The network has the ability to identify contemporaneous (zero-lag) causal relationships between traffic flows at different locations, however our causal connections are generally subject to a temporal lag.

Working with GPS data from taxis, Myrovali et al. examined time series of traversal times for 10 roads in a region of Thessaloniki, Greece~\cite{myrovali2021spatio}. They applied multivariate Granger causality analysis from a MATLAB toolbox. They were able to intuitively explain certain edges from the resulting causal network representing causal connections among the 10 road segments.

In the work by Jung et. al~\cite{jung2024catom}, the authors build an interactive system to compute and display causality connections between road locations computed with the Granger causal density test of Seth et al.~\cite{seth2005causal}. Tests with domain experts showed the system to be qualitatively effective for exploring road traffic causality, although no quantified accuracy results were given.

Mao et al. present a state space solution for discovering traffic causality from time series of speed measurements~\cite{mao2025convergent}. The state space for a sensor consists of its speed values at different time lags, and an analysis produces a full causality strength matrix for the whole set of sensors. The results are shown to be reasonable on two sensors and their nearby neighboring sensors. Our method, in contrast, is tested on hundreds of ground truth sensor pairs, and we employ a machine learning approach to improve accuracy.

\section{Traffic Data} \label{sec:data}
\begin{figure}[ht]
    \centering
    \includegraphics[width=1\linewidth]{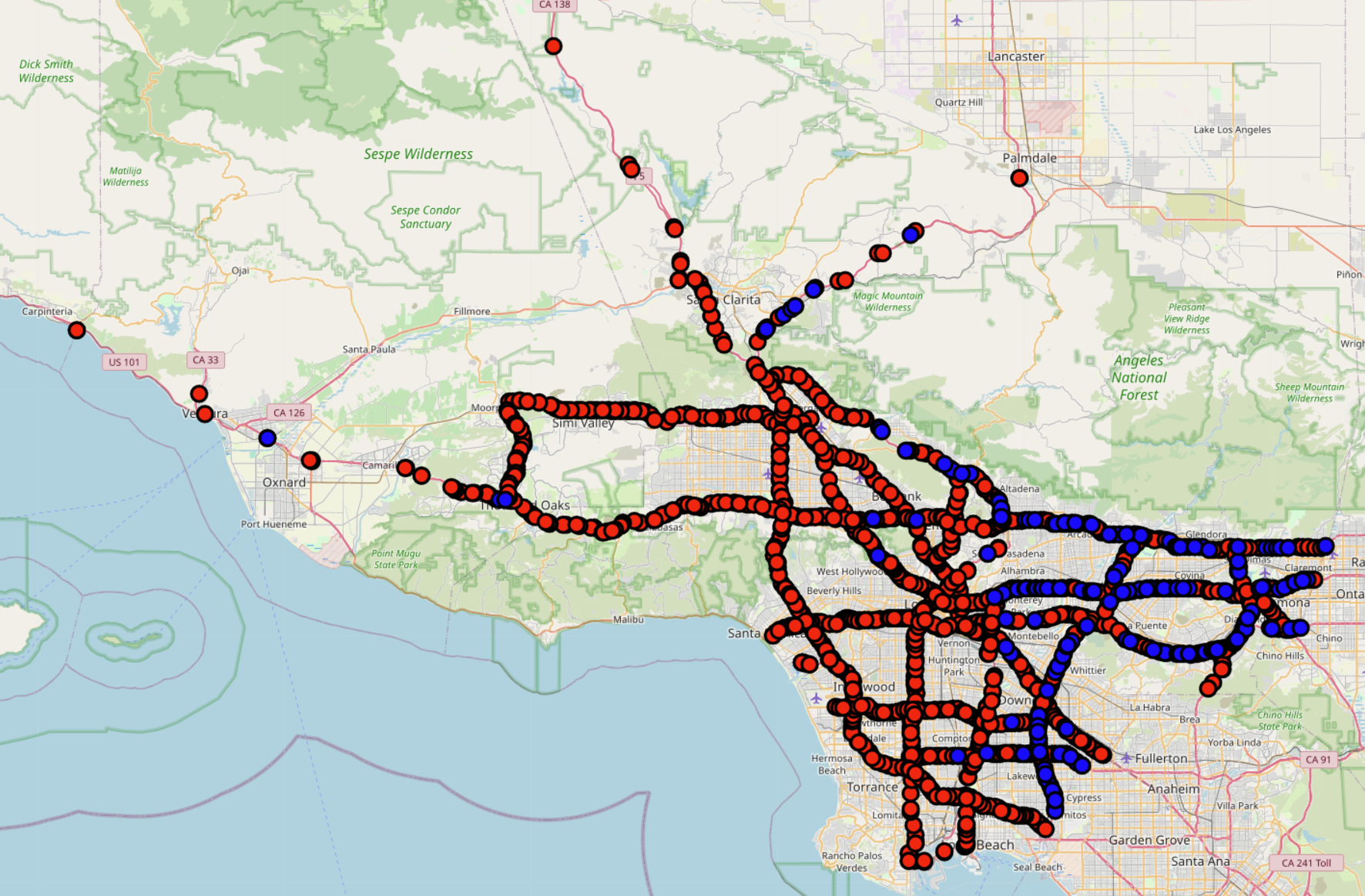}
    \caption{Locations of ``Mainline'' sensors in Caltrans District 7. The blue dots represent the 195 stations we used in our causality analysis, because they had sufficient speed data.}
    \label{fig:station_map}
\end{figure}

Our traffic data comes from California's Caltrans Performance Measurement System (PeMS)~\cite{pems_dot_ca_gov}, which provides free downloads of traffic data from static sensors on the freeway system across all major metropolitan areas of the state. Our data was from the Los Angeles area (District 7) for the first six months of 2024. Some of the measurement station locations are shown in Figure~\ref{fig:station_map}. We used the ``Station - 5-Minute'' data, which gives, for each measurement station, timestamped five-minute aggregates of mean speed and flow for each traffic lane. Our analysis was based on mean traffic speed over all the lanes for each station, maintaining the five-minute measurement intervals for maximum temporal resolution. Typical speed data for one station for one week is shown in Figure~\ref{fig:traffic_slowdown_events}.



\begin{figure}[ht]
    \centering
    \includegraphics[width=1\linewidth]{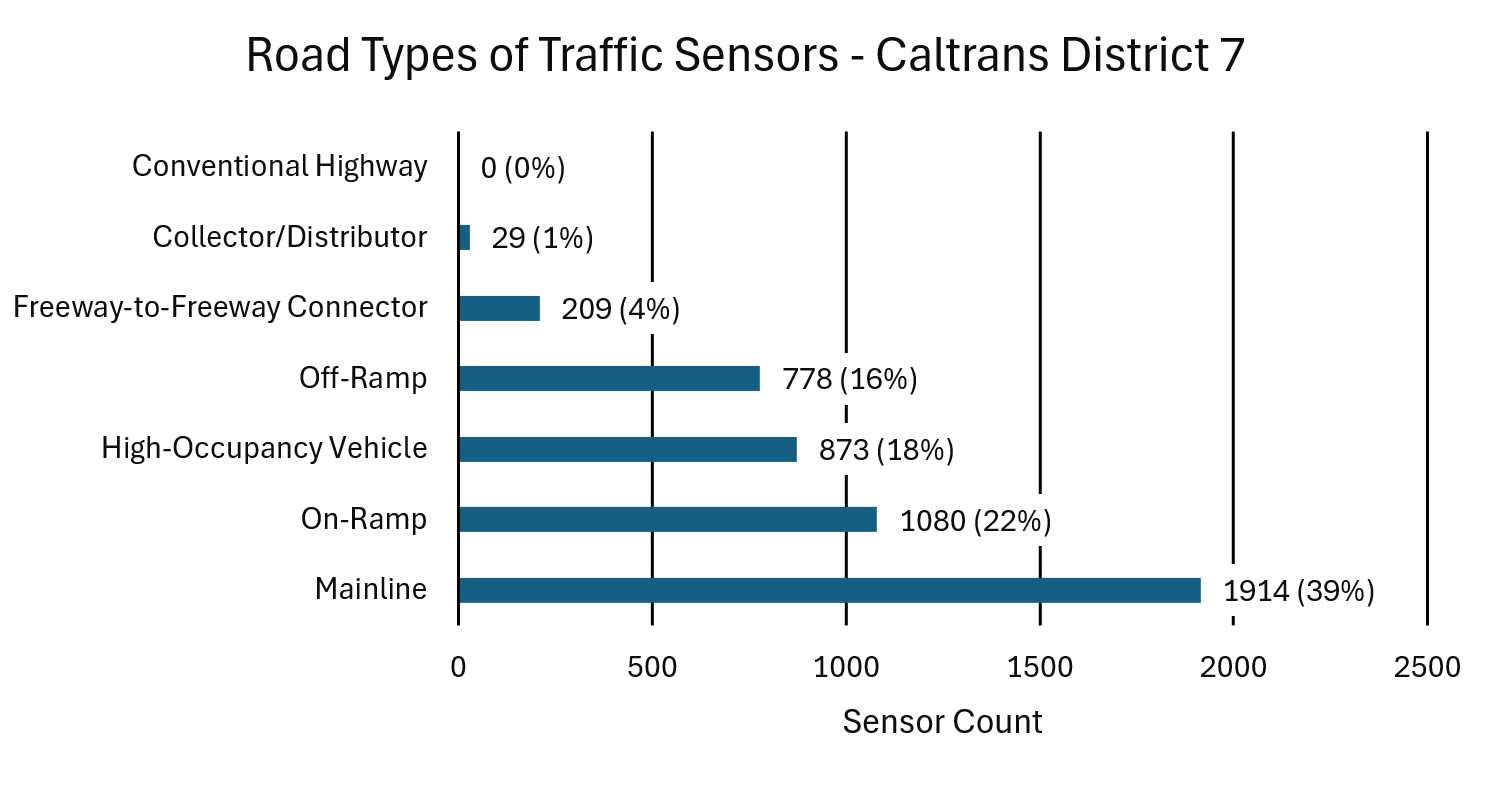}
    \caption{Counts of sensor road types in Caltrans District 7}
    \label{fig:station_counts}
\end{figure}

The sensor types and the total number of each in District 7 are shown in Figure~\ref{fig:station_counts}. We chose only sensors on the ``Mainline'' roads to reduce the computational load in our experiments and because these roads give clear instances of very likely causal connections for testing. The ``Mainline'' sensors are shown in Figure~\ref{fig:station_map}.

Many of the speed values in the PeMS data are imputed due to sensor dropouts~\cite{PeMS_Intro_User_Guide_v6}. The imputation methods do not necessarily preserve the anomalous traffic slowdowns on which our algorithm depends, so stations with much imputed data are not good candidates for our causality analysis. Figure~\ref{fig:station_data_completeness} shows how the number of sensor stations available varies with different levels of data completeness, where completeness is the fraction of non-imputed data. The figure shows, for instance, that 300 stations have 75\% or more non-imputed data. For our analysis, we chose 195 stations with at least 90\% complete data, shown as blue dots in Figure~\ref{fig:station_map}. 

\subsection{Drive Time Data}
For training and testing, we select ordered pairs of stations that very likely have a causal connection or very likely do not have a causal connection, as explained next in Section~\ref{subsec:ground_truth}. Part of this ground truth selection depends on the free-flow driving time between pairs of stations. We accomplish this by constructing a drive time matrix using the Bing Maps API~\cite{bingmapsapi}. For each station $i$, we query the API to provide us with the time it takes to travel in a motorized vehicle to each other station $j$ such that $j \neq i$. Each entry $D_{ij}$ of this drive time matrix matrix $D$ represents the time it takes to drive from station $i$ to station $j$. It is seen trivially that $D_{ii} = 0$ for all $i$. Because the measured drive time from the API was dependent on traffic at that time, we selected an arbitrary time (i.e. March 3rd, 2024 at midnight). This helps ensure that congestion does not play a strong role in the computed travel time between stations, because we are most interested in the free-flow drive times for selecting ground truth pairs.

\subsection{Ground Truth} \label{subsec:ground_truth}
Our ultimate goal is to identify cause/effect station pairs, indicating that traffic at the ``cause'' station affects traffic at the ``effect'' station. We select these pairs from the 195 measurement stations that had enough non-imputed data as described above. In selecting ground truth pairs, we consider all candidate causal pairs $(i,j,k)$, where $i$ is the potential causal station, $j$ is the potential affected station, and $k$ is the temporal lag from cause to effect. The lag $k$ is an integer multiple of five minutes, because the time series is sampled every five minutes. This set contains $^NP_2l_{max} = N(N-1)l_{max}$ unique edges, where $l_{max}$ represents the total number of lags considered and $N=195$ is the total number of stations considered. 


\begin{figure}[ht]
    \centering
    \includegraphics[width=1\linewidth]{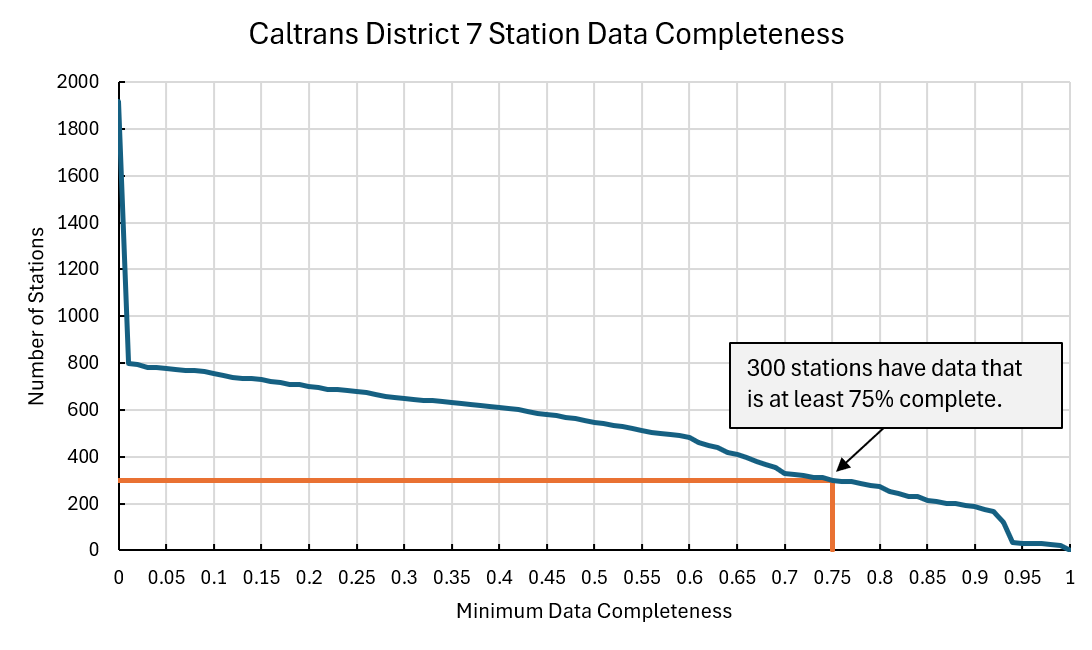}
    \caption{Counts of Mainline sensors vs data completeness in Caltrans District 7}
    \label{fig:station_data_completeness}
\end{figure}


From this set of candidate causal pairs, we select ground truth pairs as positive (causal) or negative based on several criteria. We attempt to choose pairs where the causality is very clear, and thus we are conservative in our selections in order to create an accurate ground truth set. We are essentially looking for obvious causal pairs like the one illustrated in Figure~\ref{fig:ground_truth_positive_example}. Our criteria for selecting positive ground truth pairs are as follows:

\noindent \textbf{Distance required to travel}: We classify pairs of stations $(i,j)$ as negative if their drive time is too long. Based on prior work, we posit that the maximum speed for congestion to propagate in the upstream direction of travel from a causal point is roughly 20 kph~\cite{fei2017practical}. Given this and the distance to travel, we can compute the approximate time required for traffic to propagate upstream to the affected station. This distance is given to us by the drive time matrix $D$ described above. This approach allows us to compute the exact \textit{minimum} lag that we expect a cause to propagate. This is a minimum, because a slowdown could take longer to propagate than the minimum time. For example, if stations $i$ and $j$ are ten kilometers apart, we say that it should take approximately 30 minutes (or a lag of six) to propagate to the affected station. For simplicity, we allow a soft threshold of one, meaning in our previous example, we consider lags of six and seven to be fair estimates of the propagation time. In this case, lags zero to five and lag eight for the two stations would be labeled negative, and six and seven would be positive. A lag of zero is considered non-causal since we assume propagation cannot be instantaneous. In our experiments, we set the maximum lag $l_{max} = 8$ (40 minutes) and do not consider any stations beyond this threshold.

\noindent \textbf{Road Number and Direction}: An important consideration is how we consider stations on different major highways. For classifying ground truth pairs, we work under the assumption that the majority of traffic events are contained to a specific road and direction (i.e. CA-110 North or I-105 East). If two stations belong to different roads and directions, we label the pair as negative for all lags. 

    
\noindent \textbf{Direction of Travel}: This refers to the direction of travel (i.e. northbound or southbound) on the \textbf{same} highway. All major highways typically offer two directions to travel on the same road (i.e. N/S or E/W). Thus, we say that if traffic is flowing in one direction, and there is a causal event, it should not impact any of the traffic flowing in the other direction. For example, traffic events on 110 North should not affect events on 110 West. In popular culture, a cause/effect relationship between two different highway directions is typically referred to as ``rubbernecking,'' but is not something we can capture easily with our data, so we eliminate these stations from consideration as positive ground truth pairs, again being conservative.
    
\noindent \textbf{Propagation Direction}: Naturally, traffic events that occur at any given station should only propagate upstream, opposite to the direction of travel. It may seem difficult initially to determine which direction is upstream from two given stations (since roads are not perfect and can curve in many unpredictable directions), but there is a simple yet intuitive approach to resolve this issue. Given two stations $i$ and $j$, we can obtain the distance to travel from $i$ to $j$ and the distance to travel from $j$ to $i$. Call these distances $D_{ij}$ and $D_{ji}$. If $D_{ij} < D_{ji}$, then we say that traffic flows naturally from $i$ to $j$, and traffic events at $j$ propagate to $i$ (and vice versa if $D_{ij} > D_{ji}$). Note that having a longer distance does not immediately classify a candidate station pair as negative, since they could still be on different major lines.

\begin{figure*}[ht]
    \centering
    \begin{subfigure}[t]{0.49\textwidth}
        \includegraphics[width=\linewidth]{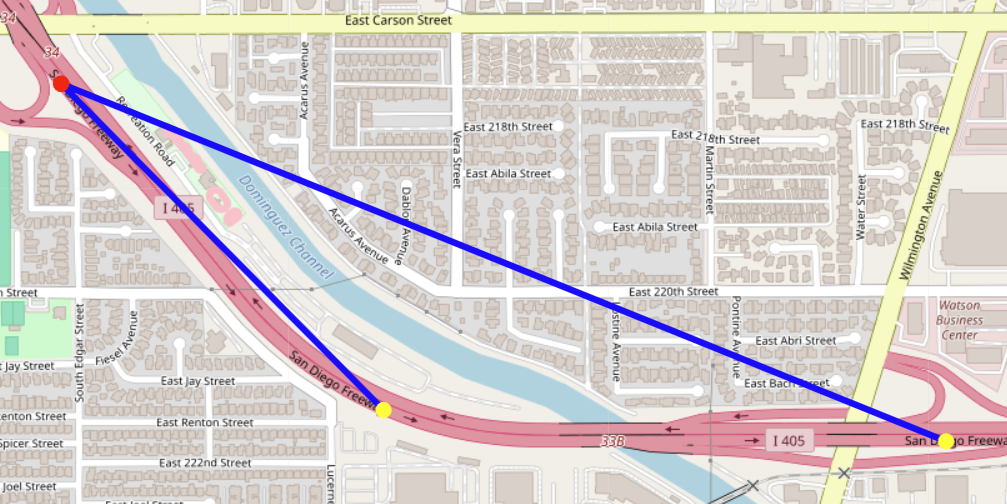} 
        \caption{Positive ground truth causal pairs.}
        \label{fig:ground_truth_map_positive}
    \end{subfigure}
    \hfill
    \begin{subfigure}[t]{0.49\textwidth}
        \includegraphics[width=\linewidth]{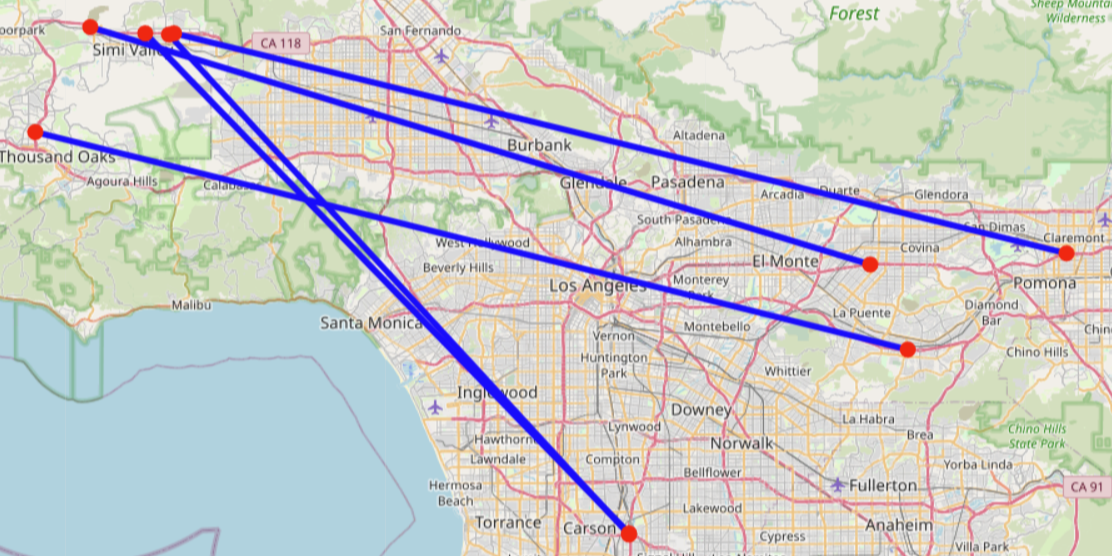}
        \caption{Negative ground truth causal pairs.}
        \label{fig:ground_truth_map_negative}
    \end{subfigure}
    \caption{Example ground truth causal pairs. In (a), yellow indicates the location of the cause (downstream from effect), and red indicates the location of effect (upstream from cause). In (b), both causal directions are considered as negative causal pairs.}
    \label{fig:ground_truth_maps}
\end{figure*}

All edges that meet these criteria are labeled positive in our dataset. For example, in Figure \ref{fig:ground_truth_map_positive}, we see nearby stations where the causal station $i$ is upstream from the affected station $j$, which is correspondingly labeled as a positive ground truth pair in our dataset. It is important when using the driving time between two stations to construct the ground truth, we use the drive time from the affected station $j$ to the causal station $i$. This gives us context as to how long it would theoretically take us to see traffic from $i$ appear at $j$. The remaining edges that do not meet this criteria are placed in a pool of possible negative edges, meaning they may not have a causal connection. We sort these edges in descending order by the driving time between them, $D_{ij}$, such that the edges with the longest driving times are near the top of the list. These are the edges with the least chance of a causal connection, so our negative pairs are chosen very conservatively. When we build a set of ground truth edges, both positive (causal) and negative (non-causal), we take the entire set of positive edges based on the criteria above, and then we take an equal number of negative edges from the sorted list of edges that are far apart in terms of driving time, giving a balanced training set. In the balanced ground truth set, the minimum drive time of any negative edge is over 75 minutes, meaning we can be fairly certain that none of the negative edges represent a causal connection. Some of these negative ground truth pairs are shown in Figure~\ref{fig:ground_truth_map_negative}.

Using these criteria, and a set of $195$ stations with $l_{max}=8$, we consider $195 * 194 * 8 = 302,640$ total candidate edge/lag tuples, from which we label $1771$ as positive ground truth edges (roughly $0.59\%$ of the total possible edge/lag tuples).

\section{Event-Based Traffic Causality} \label{sec:method}
Existing methods for detecting causality in time series tend to process the time series values directly. Such is the case with PCMCI~\cite{runge2020discovering} and DYNOTEARS~\cite{pamfil2020dynotears}, with which we compare to our method in Section~\ref{sec:results}. This leads to intense, slow processing. In contrast, our method is based on a binary time series, where each element in the time series represents whether an event occurred or not. For traffic, the event represents the beginning of a significant and unexpected reduction in traffic speed. For example, a value of 1 at a specified time and station indicates an unexpected, and 0 otherwise. These unexpected slowdowns represent natural experiments whose effects we can search for in other parts of the road network.

This section describes our algorithm, beginning with how we detect unexpected slowdowns.

\subsection{Detecting Traffic Slowdown Events}
We will introduce our mathematical notation to explain our process for finding events in the speed time series. The speed measurement of station $i$ at time $j$ is $s_i(t_j)$ for $i \in [1,N]$ and $j \in [1,M]$. Here $N=195$ measurement stations and $M=52,416$ five-minute intervals in the first six months of 2024. As described above, each station $i$ reports the average speed over all lanes every five minutes, thus $t_j$ represents a discrete time that is a whole multiple of five minutes.

To find unexpected slowdown events at a station, we must have a notion of normal traffic speed at the station. There are several ways to do this, and they all attempt to predict what the normal traffic speed will be. We experimented with different approaches to predicting traffic speed based on historical traffic speed data. It is important that the prediction not be too good, because a 100\% accurate prediction would discover nothing unexpected.

We explored two deep models for traffic speed prediction: a simple multilayer perceptron (MLP) and Long Short-Term Memory (LSTM). They both proved promising in our initial tests, but their computation time was excessive considering our six months of speed data for almost 200 traffic measurement stations to analyze. We instead chose a fast, effective method based on medians.

\begin{figure}
    \centering
    \includegraphics[width=1\linewidth]{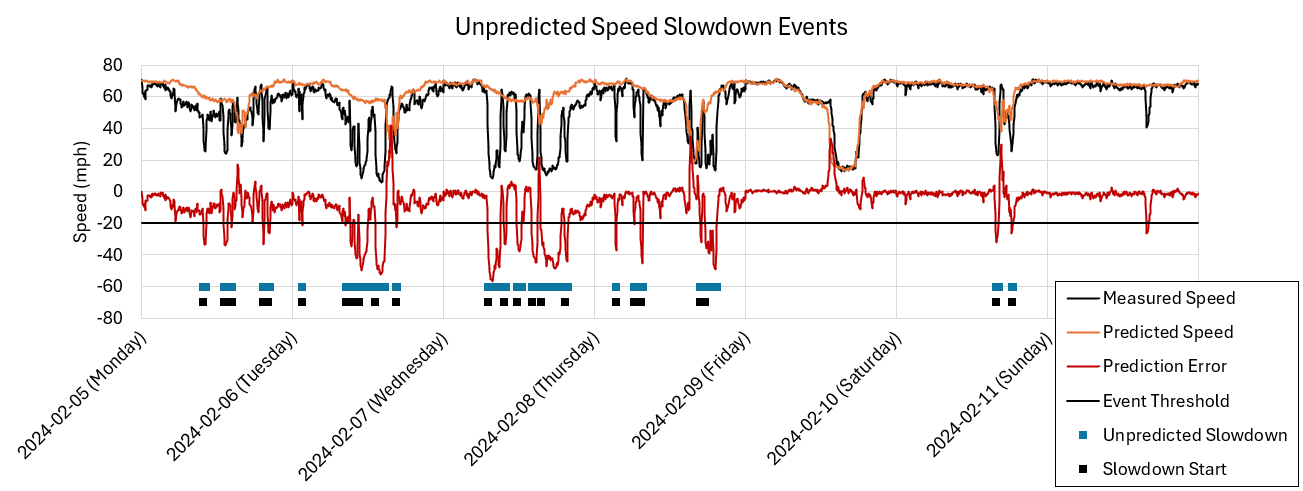}
    \caption{Traffic slowdown events are the leading edges of unpredicted slowdowns. This plot shows one week of speed data from one measurement station and its associated slowdown events.}
    \label{fig:traffic_slowdown_events}
\end{figure}

For a given measurement station, the median approach computes the median speed for each time slot in a generic week as its speed prediction. For example, the predicted speed at 10 am on a Monday is the median of all the speeds observed at that station at 10 am on all Mondays. Figure~\ref{fig:traffic_slowdown_events} shows one week of traffic speed data for one station. The black curve shows the measured speeds, and the orange curve shows the median week computed over all six months of data. This method tends to capture the normal variations in speed, ignoring unusual variations by using the median.

The predicted traffic speed is $\hat{s_i}(t_j)$, and the prediction error is $s_i(t_j) - \hat{s_i}(t_j)$. This is shown as the red curve in Figure~\ref{fig:traffic_slowdown_events}. The prediction error is negative when the actual traffic speed is slower than the prediction, indicating an unexpected slowdown. We declare a traffic slowdown when the prediction error fraction is less than a threshold $\alpha$, i.e. when $\frac{s_i(t_j) - \hat{s_i}(t_j)}{\hat{s_i}(t_j)} < -\alpha$. We experimented with different values of $\alpha$ in Section  ~\ref{subsec:ablation_study}).

Comparing the prediction error to the threshold $\alpha$ gives a binary time series indicating the presence of an unexpected traffic slowdown, i.e.

\begin{equation*}
    u_i(t_j) = 
    \begin{cases}
        \text{true} & \text{if } \frac{s_i(t_j) - \hat{s_i}(t_j)}{\hat{s_i}(t_j)} < -\alpha \\
        \text{false} & \text{otherwise}
    \end{cases}
\end{equation*}

\begin{figure}
    \centering
    \includegraphics[width=1\linewidth]{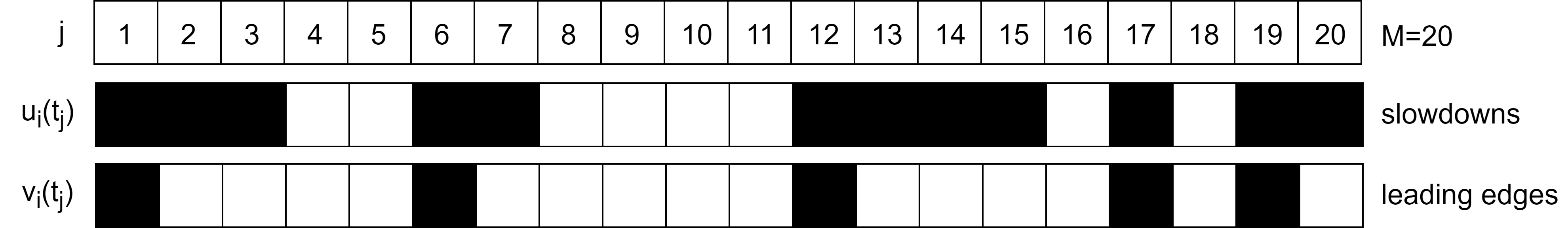}
    \caption{Black sequences of slowdowns are converted to leading edges to mark discrete events.}
    \label{fig:sequences_to_leading_edges}
\end{figure}

\noindent An example of $u_i(t_j)$ is shown as blue squares in Figure~\ref{fig:traffic_slowdown_events}, and a close-up example is shown as the top signal in Figure~\ref{fig:sequences_to_leading_edges}. We are interested in sudden speed disruptions, so we concentrate on the leading edges of traffic slowdowns indicating the start of a slowdown, notated as $v_i(t_j)$:

\begin{equation*}
    v_i(t_j) = 
    \begin{cases}
        u_i(t_j) & \text{if } j=1 \text{ (first element)}\\
        \text{true} & \text{if } u_i(t_j)=\text{true } \And u_i(t_{j-1})=\text{false} \\
        \text{false} & \text{otherwise}
    \end{cases}
\end{equation*}

\noindent An example of $v_i(t_j)$ is shown as the bottom signal in Figure~\ref{fig:sequences_to_leading_edges}. This is essentially marking the beginning of every sequence of adjacent slowdowns with a ``true''. The case for $j=1$ represents the first element.

\subsection{Probability of Cause}
We can compute the probability $p_c$ that events in one signal have caused events in another signal.

\begin{figure}
    \centering
    \includegraphics[width=1\linewidth]{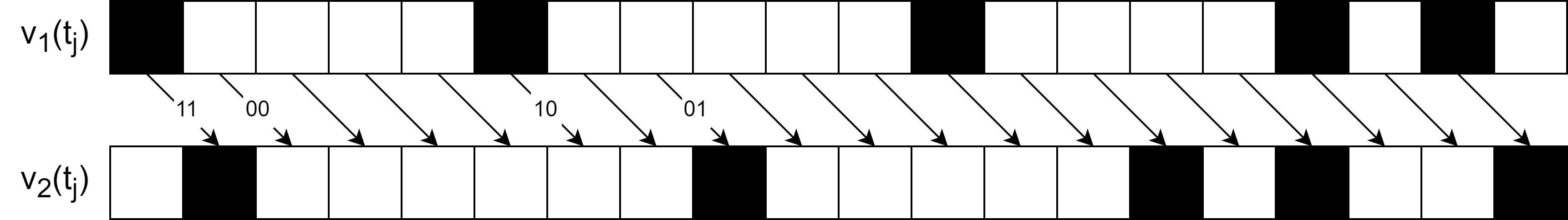}
    \caption{For a lag of one, we count the different kinds of correspondences between the hypothesized causal signal on top ($v_1(t_j)$) with the hypothesized effect signal below ($v_2(t_j)$). Examples of the four types of correspondences are marked on the arrows.}
    \label{fig:event_correspondences}
\end{figure}

As illustrated in Figure~\ref{fig:event_correspondences}, we have two binary sequences where each element of each sequence can represent the presence or absence of an event. One sequence is posited as the cause and the other as the effect, which means that events in the causal sequence sometimes cause events in the effect sequence at some constant lag in time. In Figure~\ref{fig:event_correspondences}, the supposed cause is the top signal, the supposed effect is the bottom signal, and the candidate lag is one.

In each event sequence, an event can happen spontaneously, with no apparent cause, with probability $p_s$. For traffic, example spontaneous slowdowns could come from a crash, animal crossing, or distraction. The probability that an event in the causal sequence causes an event at the corresponding time slot in the effect sequence is $p_c$. Our goal is to estimate the value of $p_c$ in order to assess the strength of the causal connection.

There are four possible scenarios for a single corresponding pair of event time slots for the cause and effect sequence, illustrated in Table~\ref{tab:paired_slots_contingency_counts}. If zero (0) represents no event and one (1) represents an event, then the four possible pairs are 00, 01, 10, and 11. 
Assuming the pair is independent of all the other pairs, the probabilities of these four possible pairs are:

\begin{itemize}
    \item \textbf{00}: No spontaneous event at the cause and no spontaneous event at the effect, thus $f_{00}(p_s,p_c) = (1-p_s)(1-p_s)$.
    \item \textbf{01}: No spontaneous event at the cause and a spontaneous event at the effect, thus $f_{01}(p_s,p_c) = (1-p_s)p_s$.
    \item \textbf{10}: A spontaneous event at the cause and no spontaneous event at the effect and no caused event at the effect, thus $f_{10}(p_s,p_c) = p_s(1-p_s)(1-p_c)$.
    \item \textbf{11}: A spontaneous event at the cause and a spontaneous event or caused event at the effect, thus $f_{11}(p_s,p_c) = p_s(p_s + p_c - p_s p_c)$.
\end{itemize}

\noindent These are illustrated in Table~\ref{tab:paired_slots_contingency_probabilities}, and we note that $f_{00}(p_s,p_c)+f_{01}(p_s,p_c)+f_{10}(p_s,p_c)+f_{11}(p_s,p_c) = 1$. Our assumption about the independence of the pairs is bolstered by the fact that we look at only the leading edges of traffic slowdown events, not the entire extent of slowdowns. Presumably, unexpected slowdowns are independent.

\begin{table}[ht]
    \centering
    \begin{tabular}{llll}
        &                        & \multicolumn{2}{c}{effect}                                    \\ \cline{3-4} 
        & \multicolumn{1}{l|}{}  & \multicolumn{1}{c|}{0}        & \multicolumn{1}{c|}{1}        \\ \cline{2-4} 
        \multicolumn{1}{l|}{\multirow{2}{*}{cause}} & \multicolumn{1}{c|}{0} & \multicolumn{1}{c|}{$A_{00}$} & \multicolumn{1}{c|}{$A_{01}$} \\ \cline{2-4} 
        \multicolumn{1}{l|}{}                       & \multicolumn{1}{c|}{1} & \multicolumn{1}{c|}{$A_{10}$} & \multicolumn{1}{c|}{$A_{11}$} \\ \cline{2-4} 
    \end{tabular}
    \caption{Paired slot counts}
\label{tab:paired_slots_contingency_counts}
\end{table}

\begin{table}[ht]
    \centering
    \begin{tabular}{llll}
        &                        & \multicolumn{2}{c}{effect}                                    \\ \cline{3-4} 
        & \multicolumn{1}{l|}{}  & \multicolumn{1}{c|}{0}        & \multicolumn{1}{c|}{1}        \\ \cline{2-4} 
        \multicolumn{1}{l|}{\multirow{2}{*}{cause}} & \multicolumn{1}{c|}{0} & \multicolumn{1}{c|}{$(1 - p_s)(1 - p_s)$} & \multicolumn{1}{c|}{$(1 - p_s)p_s$} \\ \cline{2-4} 
        \multicolumn{1}{l|}{}                       & \multicolumn{1}{c|}{1} & \multicolumn{1}{c|}{$p_s(1 - p_s)(1 - p_c)$} & \multicolumn{1}{c|}{$p_s(p_s + p_c - p_s p_c)$} \\ \cline{2-4} 
    \end{tabular}
    \caption{Paired slot probabilities}
\label{tab:paired_slots_contingency_probabilities}
\end{table}

For all the corresponding pairs in the two sequences, the counts of the different kinds of possible pairs are $A_{00}$, $A_{01}$, $A_{10}$, and $A_{11}$, as shown in Table~\ref{tab:paired_slots_contingency_counts}. Then the likelihood of all the counts over the timespan of the two segments is, assuming independence between pairs:

\begin{equation}
    L(p_s,p_c) = \prod_{i=0}^1 \prod_{j=0}^1 (f_{ij}(p_s,p_c) \big)^{A_{ij}}
\end{equation}

\noindent The log likelihood is:
\begin{align}
    \ell(p_s,p_c) =& \text{ln}\big[L(p_s,p_c) \big] \nonumber \\
    =&{A_{00}}\text{ln}\big(f_{00}(p_s,p_c) \big) + {A_{01}}\text{ln}\big(f_{01}(p_s,p_c) \big) + \nonumber \\
    & {A_{10}}\text{ln}\big(f_{10}(p_s,p_c) \big) + {A_{11}}\text{ln}\big(f_{11}(p_s,p_c) \big)
    \label{eq:negative_log_likelihood}
\end{align}

We want to find the causal probability $p_c$ (along with the spontaneous probability $p_s$) that maximizes the log likelihood. We set the partial derivatives to zero, i.e. $\frac{\partial \ell}{\partial p_s}=0$ and $\frac{\partial \ell}{\partial p_c}=0$. The partial derivatives of $\text{ln}(f_{ij})$ are

\begin{align*}
    \frac{\partial \text{ln}(f_{00})}{\partial p_s} &= \frac{2}{p_s-1} &
    \frac{\partial \text{ln}(f_{00})}{\partial p_c} &= 0 \\
    \frac{\partial \text{ln}(f_{01})}{\partial p_s} &= \frac{1-2p_s}{p_s(1-p_s)} &
    \frac{\partial \text{ln}(f_{01})}{\partial p_c} &= 0 \\
    \frac{\partial \text{ln}(f_{10})}{\partial p_s} &= \frac{1-2p_s}{p_s(1-p_s)} &
    \frac{\partial \text{ln}(f_{10})}{\partial p_c} &= \frac{1}{p_c-1} \\
    & \frac{\partial \text{ln}(f_{11})}{\partial p_s} = \frac{-2 p_s (p_c-1) + p_c}{p_s(p_s + p_c - p_s p_c)} & \\
    & \frac{\partial \text{ln}(f_{11})}{\partial p_c} = \frac{1-p_s}{p_s + p_c - p_s p_c} &
\end{align*}

\noindent Setting 

\begin{align*}
    A_{00}\frac{\partial \text{ln}(f_{00})}{\partial p_s} + A_{01}\frac{\partial \text{ln}(f_{01})}{\partial p_s} + A_{10}\frac{\partial \text{ln}(f_{10})}{\partial p_s} + A_{11}\frac{\partial \text{ln}(f_{11})}{\partial p_s} &= 0 \\
    A_{00}\frac{\partial \text{ln}(f_{00})}{\partial p_c} + A_{01}\frac{\partial \text{ln}(f_{01})}{\partial p_c} + A_{10}\frac{\partial \text{ln}(f_{10})}{\partial p_c} + A_{11}\frac{\partial \text{ln}(f_{11})}{\partial p_c} &= 0
\end{align*}

\noindent gives closed form expressions for the values of $p_s$ and $p_c$ that maximize the log likelihood, i.e.

\begin{align}
    p_s &= \frac{A_{01}+A_{10}+A_{11}}{2 (A_{00}+A_{01})+A_{10}+A_{11}} \label{eq:ps_unconstrained} \\
    p_c &= \frac{2 A_{00} A_{11}+A_{01} (A_{11}-A_{10})-A_{10}^2-A_{10} A_{11}}{(2 A_{00}+A_{01}) (A_{10}+A_{11})} \label{eq:pc_unconstrained}
\end{align}

To veryify that Equations~\ref{eq:ps_unconstrained} and~\ref{eq:pc_unconstrained} represent the location of the maximum, we first compute the determinant of the Hessian, i.e.

\begin{align*}
    D &= \frac{\partial^2 \ell(p_s,p_c)}{\partial p_s^2}\frac{\partial^2 \ell(p_s,p_c)}{\partial p_c^2} - \left(\frac{\partial^2 \ell(p_s,p_c)}{\partial p_s \partial p_c}\right)^2 \\
    &= \frac{(2 A_{00}+A_{01}) (A_{10}+A_{11})^3 (2 A_{00}+2 A_{10}+A_{10}+A_{11})}{A_{10} A_{11} (A_{01}+A_{10}+A_{11})}
\end{align*}

\noindent D is always positive, because $A_{ij} \geq 0$ and $A_{00} + A_{01} + A_{10} + A_{11} > 0$. Thus the critical point from Equations ~\ref{eq:ps_unconstrained} and~\ref{eq:pc_unconstrained} represents an extrema, not a saddle point. The critcal point is a maximum if either $\frac{\partial^2 \ell(p_s,p_c)}{\partial p_s^2} < 0$ or $\frac{\partial^2 \ell(p_s,p_c)}{\partial p_c^2} < 0$. We find that

\begin{equation*}
   \frac{\partial^2 \ell(p_s,p_c)}{\partial p_c^2} = -\frac{(2 A_{00}+A_{01})^2 (A_{10}+A_{11})^3}{A_{10} A_{11} (2 A_{00}+2 A_{01}+A_{10}+A_{11})^2}
\end{equation*}

\noindent This is negative for the same reason that $D$ is positive. Thus the critical point from Equations ~\ref{eq:ps_unconstrained} and~\ref{eq:pc_unconstrained} represents the unique maximum of the log likelihood, giving values for the probability of a spontaneous event $p_s$ and the probability of a causal connection $p_c$.

\subsubsection{Special Cases}

Since $p_s$ and $p_c$ are probabilities, we know $0 \leq p_s, p_c \leq 1$. However, $p_s$ and $p_c$ are unconstrained in Equations~\ref{eq:ps_unconstrained} and~\ref{eq:pc_unconstrained}. Since $\ell(p_s,p_c)$ is continuous, if the unconstrained maximum falls outside the constrained region, and because the only critical point of $\ell(p_s,p_c)$ is the unconstrained maximum, then the Extreme Value Theorem guarantees that the constrained maximum must occur somewhere on the edges of the constrained region. We look at the four edges next.

\vspace{0.5cm}
\noindent \textbf{Edge $0 \leq p_s \leq 1$ and $p_c = 0$}: \\
\noindent We have
\begin{align*}
    f_{00}(p_s,p_c) &= (1-p_s)(1-p_s) \\
    f_{01}(p_s,p_c) &= (1-p_s)p_s \\
    f_{10}(p_s,p_c) &= p_s(1-p_s)p_s \\
    f_{11}(p_s,p_c) &= p_s^2
\end{align*}

\noindent Substituting these expressions into the log likelihood of Equation~\ref{eq:negative_log_likelihood}, computing the derivative with respect to $p_s$, setting it to zero, and solving gives

\begin{align}
    p_{s,p_c=0} &= \frac{A01+A10+2A11}{2 (A00+A01+A10+ A11)} \label{eq:p_s_when_p_c_is_zero} \\
    p_c &= 0 \nonumber
\end{align}

\vspace{0.5cm}
\noindent \textbf{Edge $0 \leq p_s \leq 1$ and $p_c = 1$}: \\
\noindent We have
\begin{align*}
    f_{00}(p_s,p_c) &= (1-p_s)(1-p_s) \\
    f_{01}(p_s,p_c) &= (1-p_s)p_s \\
    f_{10}(p_s,p_c) &= 0 \\
    f_{11}(p_s,p_c) &= p_s
\end{align*}

\noindent Following the same steps as above gives

\begin{align}
    p_{s,p_c=1} &= \frac{A01+A11}{2 (A00+A01) + A11} \label{eq:p_s_when_p_c_is_one} \\
    p_c &= 1 \nonumber
\end{align}

\vspace{0.5cm}
\noindent \textbf{Edge $p_s = 0$ and $0 \leq p_c \leq 1$}: \\
\noindent In this case, the derivative of the log likelihood with respect to $p_c$ is zero, meaning there is no solution for the extreme value. Practically, when $p_s = 0$, there are no spontaneous events in the candidate causal signal, which means there is no opportunity to observe an event in the effect signal that was caused by the candidate causal signal. Thus it is impossible to detect causality in the case $p_s = 0$, and we can say nothing about $p_c$ in this case.

\vspace{0.5cm}
\noindent \textbf{Edge $p_s = 1$ and $0 \leq p_c \leq 1$}: \\
\noindent This is similar to the case above in that the derivative is zero with respect to $p_c$. When we have $p_s=1$, this means that every point in both signals is considered an event, and thus it is impossible to infer anything about $p_c$ in this case.

\subsubsection{Algorithm for Causality Probability}
The derivations above suggest this algorithm for computing $p_c$, the probability of causality.

\begin{enumerate}
    \item Compute $p_s$ and $p_c$ from Equations~\ref{eq:ps_unconstrained} and~\ref{eq:pc_unconstrained}, respectively. If $0 \leq p_s \leq 1$ and $0 \leq p_c \leq 1$, then take these values and stop.
    \item If $p_s$ or $p_c$ are outside the acceptable ranges, then compute candidate values of $p_s$ from Equations~\ref{eq:p_s_when_p_c_is_zero} giving $(p_s,p_c) = (p_{s,p_c=0},0)$  and~\ref{eq:p_s_when_p_c_is_one} and giving $(p_s,p_c) = (p_{s,p_c=1},1)$. Insert both these pairs into the log likelihood expression in Equation~\ref{eq:negative_log_likelihood} and take whichever pair gives the greater value.
\end{enumerate}

Furthermore, we introduce a slight modification in parametrized correspondences, a tolerance parameter called $\tau$. 
In the methodology described in this section, $\tau=0$.
By increasing the value of $\tau$, we allow the $A_{11}$ correspondence count with lag $l_i$ to consider all lags from $l_i \leq t \leq l_i + \tau$.
This way, we can loosen the restriction on exact correspondence and allow for some error in the temporal propagation time of traffic events.

\begin{table*}[h!]
    \centering
    \begin{tabular}{|l|c|c|c|c|}
        \hline
        \textbf{Framework} & \textbf{Balanced Scalar Threshold AUC} & \textbf{Balanced RF AUC} & \textbf{Full RF AUC} & \textbf{Time} \\
        \hline
        DYNOTEARS & 0.5769 & 0.5626 & 0.4939 & 111:32:06 \\
        \hline
        F-PCMCI$^*$ & 0.5003 & 0.5000 & 0.4999 & 33:44:12 \\
        \hline
        Granger & \textbf{0.7156} & 0.9185 & 0.7119 & 00:22:54 \\
        \hline

        \textbf{NEXICA $\{A_{00},A_{01},A_{10},A_{11}\}$} & - & \textbf{0.9995} & \textbf{0.8851} & \textbf{00:01:51} \\
        \hline

        \textbf{NEXICA $\{A_{00},A_{01},A_{10},A_{11}, p_c\}$} & - & 0.9984 & 0.8440 &  \textbf{00:01:51} \\
        \hline

        \textbf{NEXICA $\{p_c\}$} & 0.6533 & 0.8141 & 0.5998 & \textbf{00:01:51} \\
        \hline
        
    \end{tabular}
    \caption{
    Area Under Curve (AUC) of \textbf{NEXICA} versus existing baselines on  ground truth data. 
    Balanced Scalar Threshold AUC and Balanced RF (Random Forest) AUC use a balanced ground truth of positive ground truth and the farthest driving time negative ground truth pairs. 
    \textbf{Balanced Scalar Threshold AUC} uses the best feature (highest AUC) from each experiment as a scalar threshold, whereas \textbf{Balanced RF AUC} uses all the parameters learned from the experiment.
    \textbf{Full RF AUC} uses the entire ground truth dataset.
    Time is in \textit{HH:MM:SS} format.
    Our proposed method not only outperforms existing baselines, but is also more computationally efficient and scalable to larger datasets.
    }
    \label{tab:model_performance}
\end{table*}

\subsection{Algorithm}

Although the above approach works for a single pair of stations and set lag, we provide an approach that scales our approach to a set of stations and outputs their results. First, we provide our framework with the set of $N$ stations (nodes) that we want to consider, the maximum lag $l_{max}$ (i.e., a maximum lag of $8$ corresponds to a maximum traffic propogation of $5*8=40$ minutes), and historical time series data for each station as input. For our experiments, we used 6 months of Caltrans PeMS data described in Section \ref{sec:data}. We then use the median week approach from Section~\ref{sec:method} to determine events representing the leading edges of traffic slowdowns.

We then consider all $^NP_{2}$ candidate permutations of station pairs. For simplicity, we do not evaluate a station against itself (i.e., no pair of stations $(i, j)$ where $i=j$), but we do consider $(i, j)$ and $(j, i)$ to be distinct pairs. For each candidate pair of stations $(i, j)$, we consider all the lags $k$ where $1 \leq k \leq l_{max}$, and model the probability of a causal relationship as explained in Section~\ref{sec:method}. When we model the probability of a causal relationship, we can adjust the error threshold $\alpha$ to increase or decrease the sensitivity of classifying a period as anomalous. If a pair of stations $(i, j)$ at lag $k$ produces a causal probability $p_c$ greater than some threshold, we declare $(i, j)$ to be a causal pair, i.e. traffic slowdowns at station $i$ cause traffic slowdowns at station $j$ with a temporal lag of $k$.

As an alternative inference approach, using the features obtained for our computation of $p_c$ and labels from our ground truth data, we train a random forest on a random selection of training data.
We then validate the effectiveness of the model at classifying ground truth pairs correctly by evaluating the model's classification ability on the remainder of the data.

\subsection{Set Arithmetic}
We employ a set-theoretic approach to efficiently determine the correspondence counts between two timeseries. In our 6 months of PeMS data, we define $T$ as the set of 52540 time intervals, i.e. $T = \{1,2,...,52540\}$. Furthermore, given a causal station $i$, effect station $j$, and lag $l$, we can define $C_i$ as the set of all anomalies from causal station in $T$, $E_j$ the set of all anomalies at the effect station in $T - l$. We also define $\overline{\cdot}$ as the set complement operation. We calculate the values of $A_{ij}$ as follows:


$$A_{11} = |C_i \cap E_j|$$ 
$$A_{01} = |\overline{C_i} \cap E_j|$$
$$A_{10} = |C_i \cap \overline{E_j}|$$
$$A_{00} = |T| - A_{11} - A_{01} - A_{10} $$

The complexity of this algorithm is upper-bounded by the time complexity of the set intersection operation, which has an average time complexity of $ O(\min(|E|, |C|))$.

\section{Results} \label{sec:results}

\begin{figure*}
    \centering
    \begin{subfigure}[t]{0.33\textwidth}
        \includegraphics[width=\linewidth]{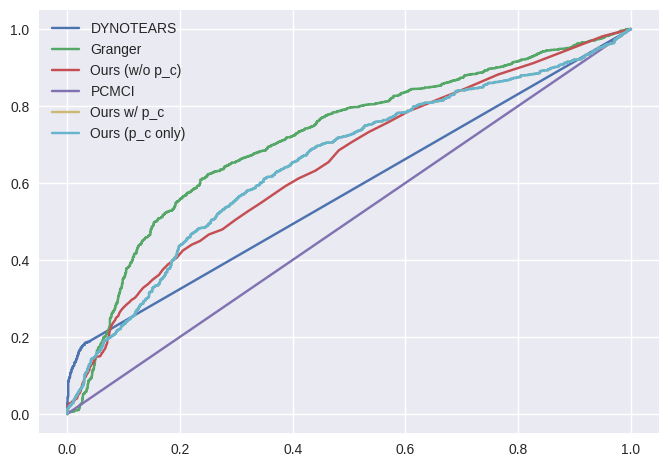} 
        \caption{Balanced Scalar Threshold ROC Curves.}
        \label{fig:balanced_roc}
    \end{subfigure}
    \begin{subfigure}[t]{0.33\textwidth}
        \includegraphics[width=\linewidth]{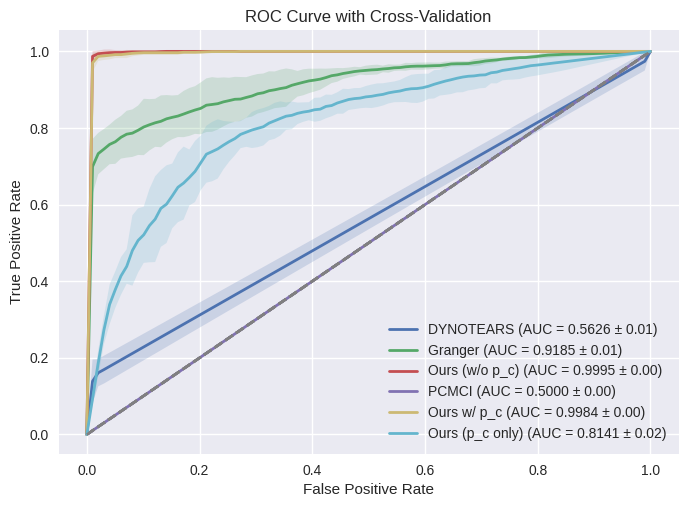} 
        \caption{Balanced Random Forest ROC Curves.}
        \label{fig:balanced_roc}
    \end{subfigure}
    \hfill
    \begin{subfigure}[t]{0.33\textwidth}
        \includegraphics[width=\linewidth]{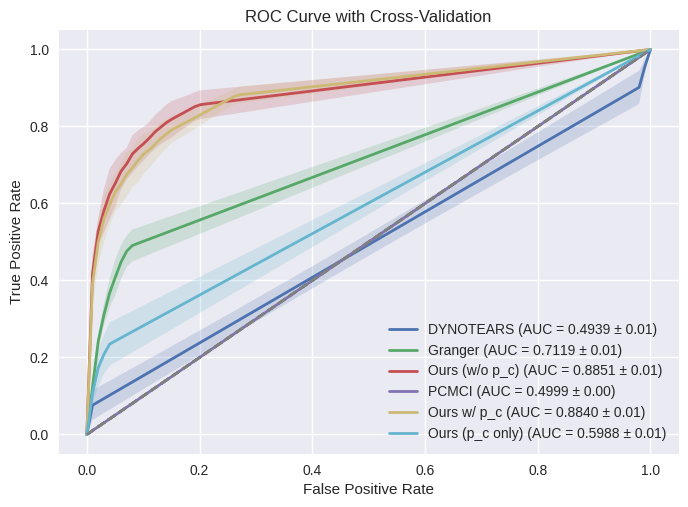}
        \caption{Full Random Forest ROC Curves.}
        \label{fig:full_roc}
    \end{subfigure}
    \caption{ROC curves for various approaches on both the balanced and full datasets against SoTA, with AUC $\pm$ standard deviation of 2 from cross validation. Data and timing is presented in Table~\ref{tab:model_performance}. For traffic causality, \textbf{NEXICA} outperforms existing methods in causality inference.}
    \label{fig:roc_curve_visualization}
\end{figure*}

We evaluated our approach on the first six months of PeMS data from 2024 on 195 select traffic sensors with at least $90\%$ complete data, as illustrated in Figure~\ref{fig:station_data_completeness}.
All variations of our experiments use $l_{max}=8$.
In preliminary experiments, we tested our approach using data without the restriction of having a high threshold of data completeness.
We found that our model struggled to properly identify anomalous periods and did not accurately classify our ground truth data.
We attribute this to the event-based nature of our approach: imputed data is by nature non-anomalous and, therefore, hides events.

We note that the size of our experiment is large compared to previous work in traffic causality discovery. For example,~\cite{queen2009intervention} looked at one road intersection, and~\cite{mao2025convergent} looked at two small clusters of road sensors.

Because the number of positive ground truth edges in our dataset is significantly less than the number of negative edges (since most station pairs and station lags are causally improbable/impossible), we not only evaluate NEXICA on the full ground truth data, but we also evaluate on a so-called "balanced" set of ground truth data.
We construct our balanced set by reducing the number of negative edges to only consider the most extreme examples, i.e. edges with the highest driving times $D_{ij}$. 
This way, we retain a $1:1$ ratio between positive and negative ground truth edges in our dataset to ensure a more realistic evaluation of our results.

\subsection{Comparison to State-of-the-Art Baselines}

Table \ref{tab:model_performance} shows the performance of our algorithm versus state-of-the-art benchmarks PCMCI and DYNOTEARS described in Section~\ref{sec:related_work}. 
For efficiency and fair comparison, we conducted all of our experiments on an NVIDIA GeForce RTX 3090 and AMD Ryzen 9 5900 12-Core Processor.
For the PCMCI baseline, we found it computationally much too slow to run using the entire timeseries, so we ran with a fraction of it (1 week) using the same 195 stations.
Furthermore, we ran an improved version of the PCMCI algorithm, F-PCMCI~\citep{castri2023fpcmci} for faster and more robust inference.

We learn a complex decision boundary for each candidate method by using a random forest of decision trees.
Each random forest uses 1000 estimators and leverages the ground truth data as labels during training and testing.
We then extract a classification probability $p_{forest}$ for each candidate tuple $(i,j,k)$, determined from the classification vote in the ensemble of trees. 
We train and test each forest on the balanced and full ground truth data, using cross-validation with five folds.

For decision tree features, we experiment with the correspondence counts $A_{00}$, $A_{01}$, $A_{10}$, and $A_{11}$ and $p_c$ (from the MLE equation in Section~\ref{sec:method}) for our method.
Our baselines use the following computed features: DYNOTEARS computes the weight of the edge in the final DAG; F-PCMCI computes the weight of the edge in the final graph along with its significance (p-value); Granger causality computes the values of a Params F test, SSR F test, SSR Chi Squared test, and a Likelihood-ratio test.

We present the results of NEXICA versus the existing state-of-the-art baselines in Table \ref{tab:model_performance}.
In the Balanced Scalar Threshold Evaluation, Granger causality outperforms our method.
We find that scalar thresholding on $p_c$ is the second most effective, and omit results for thresholding on $A_{ij}$, since thresholding on a single correspondence count is not well-defined within our framework (we experimented with this, and found that $p_c$ outperformed all other correspondence counts by scalar thresholding).

However, using all four correspondence counts ($A_{00}$ etc.) with decision trees enables NEXICA to outperform all SoTA baselines in both the balanced and full sets.
Our best classifier used the correspondence counts and omitted the computed $p_c$ as features, and achieved near-perfect accuracy when evaluated with cross validation.
We hypothesize that the information contained in the raw correspondence counts was more informative and subtle than the single causal probability value $p_c$, which resulted in a higher classification accuracy during evaluation.

We use a careful process to choose positive ground truth pairs, i.e. pairs where we are confident of a causal connection (Section \ref{sec:data}).
The remaining pairs could be causal or not.
In our balanced ground truth dataset, we sort the remaining pairs to choose only those with the largest driving times, hypothesizing that distant points on the road are not causally connected, or at least very weakly.
As mentioned previously, the negative pairs in the balanced dataset are separated by at least 75 minutes of driving. The unbalanced dataset uses all the remaining pairs as negative ground truth, potentially including pairs that may actually be causal.
Thus, we experimented with gradually, conservatively unbalancing the dataset, adding more negative ground truth pairs by taking increasingly nearer pairs from the remaining set.
Appendix~\ref{sec:unbalanced Set Ablations} contains results (Table~\ref{tab:model_performance_ratio_test}) that show that NEXICA is robust to unbalanced training sets, maintaining a high AUC and outperforming the baselines.
Appendix~\ref{sec:unbalanced Set Ablations} also contains the ROC curves along with the minimum driving times of the negative ground truth pairs in each unbalanced ground truth set.

In Figure \ref{fig:roc_curve_visualization}, we plot the ROC curves of all 3 methods from Table~\ref{tab:model_performance}. 
The visually peculiar behavior from the baselines DYNOTEARS and F-PCMCI are because these approaches yield a nonzero weight for a very small number of edges in the causal graph.
As a result, the binary classifier struggles to learn an appropriate model from these algorithms' features.

\subsection{Ablation Study}\label{subsec:ablation_study}

We provide a brief ablation study for our algorithm, investigating the impact of several hyperparameters during the learning process and the significance of each correspondence count during the classifier training.

\begin{table}[h]
    \centering
    \begin{tabular}{|l|c|c|c|c|c|}
        \hline
        \textbf{$\alpha$} & \textbf{$\tau$} &  \textbf{Balanced AUC} & \textbf{Full AUC} & \textbf{Time (s)} \\
        \hline
        0.05 & 0 & 0.9960 & 0.8003 & 264 \\
        \hline
        0.1  & 0 & 0.9976 & 0.8241 & 161 \\
        \hline
        0.15 & 0 & 0.9974 & 0.8353 & 127 \\
        \hline
        0.2  & 0 & 0.9976 & 0.8595 & 110 \\
        \hline
        0.25 & 0 & \textbf{0.9996} & 0.8590 & \textbf{98}  \\
        \hline
        0.05 & 1 & 0.9975 & 0.8422 & 304 \\
        \hline
        0.1  & 1 & 0.9984 & 0.8592 & 194 \\
        \hline
        0.15 & 1 & 0.9978 & 0.8635 & 153 \\
        \hline
        0.2  & 1 & 0.9969 & 0.8827 & 128 \\
        \hline
        0.25 & 1 & 0.9995 & \textbf{0.8851} & 111 \\
        \hline
    \end{tabular}
    \caption{Grid search of different levels of anomaly threshold $\alpha$ and tolerance $\tau$ for the corresponding results using 195 stations. Averages obtained from cross-validation. }
    \label{tab:grid_search_alpha}
\end{table}

\noindent \textbf{Grid Search.} In Table~\ref{tab:grid_search_alpha}, we present a grid search of different values of $\alpha$ (the relative threshold for declaring a traffic slowdown) and $\tau$ (the tolerance parameter for selecting a correspondence).
By increasing the alpha threshold, we effectively tell our model to be more conservative about what regions of the time series are classified as anomalous, thus computing a smaller set of natural events. 
Because increasing $\alpha$ decreases the number of intervals classified as anomalous, this decreases the execution time of our algorithm. 
Similarly, a higher $\tau$ leads to more correspondences selected, increasing the runtime of our algorithm.
We discover that ($\alpha=0.25$, $\tau=0$) are the most effective hyperparameters for correctly classifying our ground-truth pairs for the balanced dataset, and ($\alpha=0.25$, $\tau=1$) the most effective for the full dataset.

\begin{table}[h]
    \centering
    \begin{tabular}{|c|c|c|c|c|}
    \hline
    $A_{00}$ & $A_{01}$ & $A_{10}$ & $A_{11}$ & \textbf{Balanced AUC} \\
    \hline
    Yes & No  & No  & No  & $0.7871$ \\
    No  & Yes & No  & No  & $0.7975$ \\
    No & No & Yes  & No  & \textbf{0.8703} \\
    No  & No  & No & Yes  & $0.7953$ \\

    \hline

    Yes & Yes  & No & No  & $0.9885$ \\
    No  & Yes & Yes & No  & \textbf{0.9920} \\
    No & No & Yes & Yes  & 0.9786 \\
    Yes  & No  & No  & Yes & 0.9076 \\
    Yes & No  & Yes  & No & 0.9889 \\
    No  & Yes & No  & Yes & 0.9562 \\

    \hline

    Yes & Yes & Yes  & No & 0.9961 \\
    Yes  & Yes  & No & Yes & 0.9928 \\
    Yes & No  & Yes & Yes & \textbf{0.9974} \\
    No  & Yes & Yes & Yes & 0.9987 \\

    \hline

    Yes & Yes & Yes & Yes & \textbf{0.9995} \\
    \hline
    \end{tabular}
    \caption{Balanced AUC for Different Combinations of Correspondence Counts $A_{ij}$. Categories are separated by the number of ``Yes''.}
    \label{tab:aij_ablation}
\end{table}

\noindent \textbf{Effect of $A_{ij}$ on Classifier Accuracy.}
To understand the impact of each $A_{ij}$ on the accuracy of our binary classifier, we perform an ablation on all $2^4-1=15$ combinations of including and excluding $A_{ij}$ from training the random forest (omitting the one without using any of the features) and present the results in Table \ref{tab:aij_ablation}.
We discover that the most powerful contributors to the classification accuracy are $A_{01}$ and $A_{10}$, with an AUC of $99.2\%$.
We found that the pair of values $A_{01}$ and $A_{10}$ are the least correlated output values from the algorithm $(r=-0.07)$.
We hypothesize that the independent information presented in $A_{01}$ and $A_{10}$ makes their joint contribution to the classification accuracy especially high.
Furthermore, we note that since there are exactly 52540 time intervals for each timeseries in our data, knowing the value of three of the parameters is enough to determine the value of the fourth, which is illustrated by the effectiveness of the ablations omitting a single feature.

\section{Conclusion}
\label{sec:conclusion}
Our goal is to discover which locations in the road network tend to cause slowdowns at other locations. These especially causal locations are good candidates for further investigation and investment. Time series of speed data serve as a foundation for causal discovery. This problem stands out as one that offers somewhat certain ground-truth data in that certain pairs of road locations clearly have a cause/effect relationship, and certain pairs do not. This means we can not only quantify the accuracy of our results, but we can employ a machine learning approach with ground-truth data allocated to training and testing. We found that traditional Granger causality and two state-of-the-art causality discovery methods perform poorly. The state-of-the-art methods also run very slowly on our relatively large dataset. 

Our new approach, NEXICA, uses natural traffic slowdown events detected from traffic speed time series along with a maximum likelihood algorithm that computes the probability of one sequence of events causing another.
Our tests show that it is both quick to run and accurate compared to the state-of-the-art approaches. 
Overall, we identify the following advantages of our new approach:

\noindent \textbf{Functionally Agnostic.} Our approach makes no assumption about the functional relationship between speed values on different parts of the road. For instance, Granger causality assumes a linear relationship. Instead, our approach looks only for corresponding slowdowns.

\noindent \textbf{Ignores Periodicities.} Our approach implicitly eliminates the confounding effects of time. We know that traffic varies periodically over the day and week. These periodic fluctuations can be confused as causal connections. However, we eliminate the time dependence by extracting slowdowns that are \emph{not} well-predicted by time, but are instead abnormal given the time of day and day of week.

\noindent \textbf{Event Based.} Related to the point above, our approach relies explicitly on natural experiments in the data in the form of unexpected traffic slowdowns. These events are useful for identifying how traffic slowdowns propagate from one part of the road system to other parts. By explicitly concentrating on events, we relieve the downstream causal inference system of having to discover the natural experiments.

\noindent \textbf{Reduced Data.} The time series signals are reduced from real (speed) values to binary events. This means the subsequent algorithm (Section~\ref{sec:method}) and processing are relatively fast, especially compared to the state-of-the-art baselines we tested against.

\noindent \textbf{Faster and More Accurate.} Our approach is both faster and more accurate than existing state-of-the-art baselines.

For future work, we envision the following improvements:
\begin{itemize}
    \item Expand to other regions and other road types.
    \item Experiment with different traffic event detectors.
    \item Use actual traffic incident data to find spontaneous slowdowns.
    \item Use cell phone GPS data instead of road sensor data.   
    \item Expand our mathematical algorithm to account for a causal graph with multiple causal edges and multiple signal nodes.
\end{itemize}

\footnotesize
\bibliographystyle{ACM-Reference-Format}

\normalsize

\newpage
\appendix

\section{Unbalanced Set Ablations}
\label{sec:unbalanced Set Ablations}

\begin{figure*}[t]
    \centering
    \begin{subfigure}[t]{0.49\textwidth}
        \includegraphics[width=\linewidth]{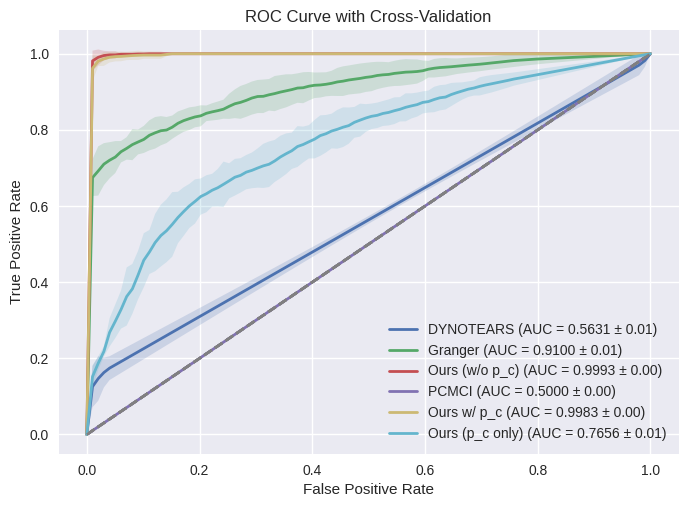} 
        \caption{1:2 positive:negative unbalanced ROC curves}
        \label{fig:balanced_roc_2}
    \end{subfigure}
    \begin{subfigure}[t]{0.49\textwidth}
        \includegraphics[width=\linewidth]{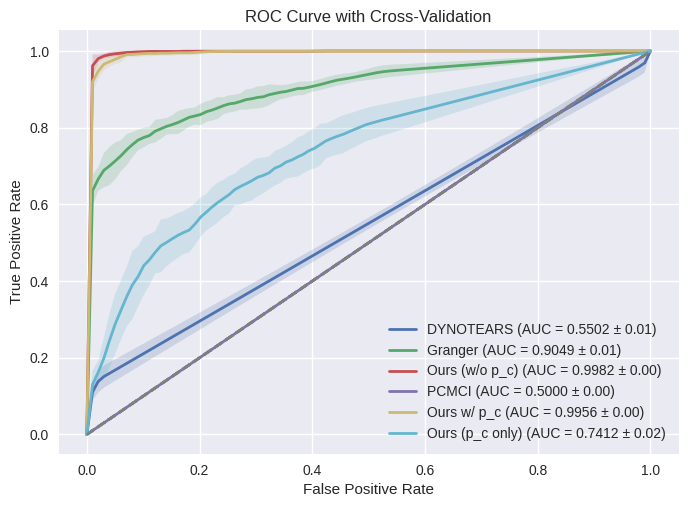} 
        \caption{1:5 positive:negative unbalanced ROC curves}
        \label{fig:balanced_roc_5}
    \end{subfigure}
    \hfill
    \caption{ROC curves for controlled unbalanced training/testing}
    \label{fig:roc_curves_remaining1}
\end{figure*}

\begin{figure*}[t]
    \centering
    \begin{subfigure}[t]{0.49\textwidth}
        \includegraphics[width=\linewidth]{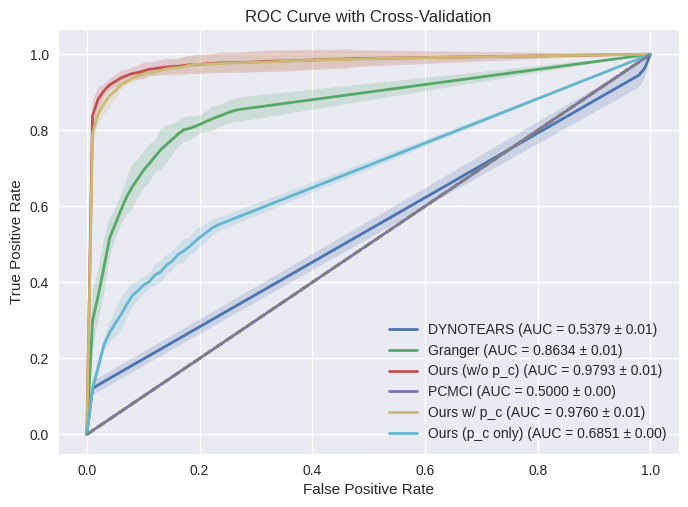} 
        \caption{1:20 positive:negative unbalanced ROC curves}
        \label{fig:balanced_roc_20}
    \end{subfigure}
    \begin{subfigure}[t]{0.49\textwidth}
        \includegraphics[width=\linewidth]{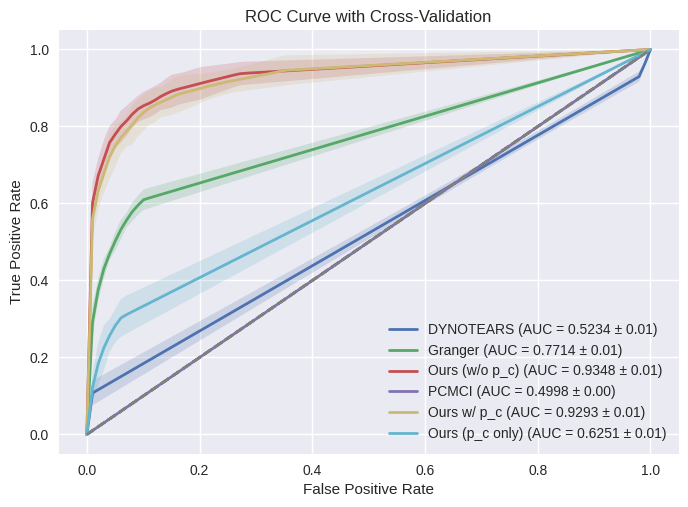} 
        \caption{1:100 positive:negative unbalanced ROC curves}
        \label{fig:balanced_roc_100}
    \end{subfigure}
    \hfill
    \caption{ROC curves for controlled unbalanced training/testing}
    \label{fig:roc_curves_remaining2}
\end{figure*}

We provide corresponding ROC curves for NEXICA and all baselines using the unbalanced $2:1$, $5:1$, $20:1$, and $100:1$ datasets in Figure~\ref{fig:balanced_roc_2}, Figure~\ref{fig:balanced_roc_5}, Figure~\ref{fig:balanced_roc_20}, and Figure~\ref{fig:balanced_roc_100}, respectively.
We also provide the corresponding AUC results in Table~\ref{tab:model_performance_ratio_test}. 
Each of these datasets also have a minimum driving time of the negative ground truth pairs of 70, 58, 37, and 18 minutes, respectively. 
We show that NEXICA is robust to various proportions of positive and clear-negative ground truth.






\begin{table*}[h]
    \centering
    \begin{tabular}{|l|c|c|c|c|c|}
        \hline
        \textbf{Framework} & \textbf{$1:1$ RF AUC} & \textbf{$1:2$ RF AUC} & \textbf{$1:5$ RF AUC} & \textbf{$1:20$ RF AUC} & \textbf{$1:100$ RF AUC} \\
        \hline

        DYNOTEARS & 0.5626 & 0.5631 & 0.5503 & 0.5379 & 0.5234 \\
        \hline

        F-PCMCI$^*$ & 0.5000 & 0.5000 & 0.5000 & 0.5000 & 0.4998 \\
        \hline

        Granger & 0.9185 & 0.9100 & 0.9049 & 0.8634 & 0.7714 \\
        \hline
        
        \textbf{NEXICA $\{A_{00},A_{01},A_{10},A_{11}\}$}
        & \textbf{0.9995} & \textbf{0.9993} & \textbf{0.9982} & \textbf{0.9793} & \textbf{0.9348} \\
        \hline

        \textbf{NEXICA $\{A_{00},A_{01},A_{10},A_{11}, p_c\}$} & 0.9984 & 0.9983 & 0.9760 & 0.9956 & 0.9293 \\
        \hline

        \textbf{NEXICA $\{p_c\}$} & 0.8141 & 0.7656 & 0.7412 & 0.6851 & 0.6251 \\
        \hline
        
    \end{tabular}
    \caption{
    Area Under Curve (AUC) of \textbf{NEXICA} versus existing baselines on datasets with varying positive-to-negative ground truth ratios. $1:X$ indicates a ratio of $X$ negative samples to $1$ positive sample. RF = Random Forest.
    }
    \label{tab:model_performance_ratio_test}
\end{table*}

\end{document}